\documentclass{article}

\usepackage[final]{corl_2017} 

\usepackage[utf8]{inputenc} 
\usepackage[T1]{fontenc}    
\usepackage{hyperref}       
\usepackage{url}            
\usepackage{booktabs}       
\usepackage[table,xcdraw]{xcolor}
\usepackage{amsfonts}       
\usepackage{nicefrac}       
\usepackage{microtype}      
\usepackage{sidecap}

\usepackage[T1]{fontenc}
\usepackage[utf8]{inputenc}

\usepackage{algorithmic}
\usepackage{chngcntr}
\usepackage[noadjust]{cite}
\usepackage{dirtytalk}
\usepackage{dsfont}
\usepackage[shortcuts]{extdash}
\usepackage{fancyhdr}
\usepackage{float}
\usepackage{hyphenat}
\usepackage{mathtools}
\usepackage{moresize}
\usepackage{newunicodechar}
\usepackage{nicefrac}
\usepackage{setspace}
\usepackage{titlesec}
\usepackage{xr}
\usepackage{xspace}
\usepackage{soul}
\usepackage{times}

\usepackage{amsmath}
\usepackage{amssymb}

\makeatletter\@ifclassloaded{ieeeconf}{}{
	\usepackage{amsthm}
}\makeatother

\usepackage{cleveref}
\usepackage[nonumberlist, nopostdot, sort=def]{glossaries}\glsdisablehyper\makenoidxglossaries
\usepackage{mathabx}
\usepackage{mleftright}
\usepackage{pdfpages}
\usepackage{thmtools}
\usepackage{zref}

\makeatletter
\ifcsname c@chapter\endcsname
\counterwithout{equation}{chapter}
\fi
\makeatother

\newcommand*{\alg}{DDCO\xspace}

\newcommand*{\slab}[1]{\label{sec:#1}}
\newcommand*{\sref}[1]{Section~\ref{sec:#1}}

\newcommand*{\C}[1]{\mathcal{#1}}

\newcommand*{\Z}[1]{\mathds{#1}}

\DeclareMathOperator{\E}{\Z E}

\DeclareMathOperator{\p}{\Z P}

\newcommand*{\copr}{\pagestyle{fancy}\thispagestyle{fancy}\renewcommand{\headrulewidth}{0pt}\lhead{}\rhead{}\rfoot{\tiny Roy Fox, \the\year}}

\newcommand*{\eq}[1]{\begin{align*}#1\end{align*}}

\newcommand*{\eqn}[1]{\begin{align}#1\end{align}}

\newcommand*{\given}{\vert}

\newcommand*{\lan}{\langle}

\newcommand*{\mrlap}{\mathrlap}

\newcommand*{\nn}{\nonumber}

\newcommand*{\ran}{\rangle}

\newcommand*{\T}{\intercal}

\renewcommand*{\vdots}{\vbox{\baselineskip=4pt\lineskiplimit=0pt\kern2pt\hbox{.}\hbox{.}\hbox{.}}}

\title{\alg: Discovery of Deep Continuous Options for Robot Learning from Demonstrations}

\author{
  Sanjay Krishnan\thanks{Equal contribution}\\
  EECS Department, UC Berkeley\\
  \texttt{sanjaykrishnan@berkeley.edu} \\
  \And
  Roy Fox$^*$\\
  EECS Department, UC Berkeley\\
  \texttt{royf@berkeley.edu} \\
  \And
  Ion Stoica\\
  EECS Department, UC Berkeley\\
  \texttt{istoica@berkeley.edu} \\
  \And
  Ken Goldberg\\
  EECS and IEOR Departments, UC Berkeley\\
  \texttt{goldberg@berkeley.edu} \\
}

\begin{document}
\maketitle


\begin{abstract}
An option is a short-term skill consisting of a control policy for a specified region of the state space, and a termination condition recognizing leaving that region.
In prior work, we proposed an algorithm called Deep Discovery of Options (DDO) to discover options to accelerate reinforcement learning in Atari games.  
This paper studies an extension to robot imitation learning, called Discovery of Deep Continuous Options (DDCO), where low-level continuous control skills parametrized by deep neural networks are learned from demonstrations.
We extend DDO with: (1) a hybrid categorical--continuous distribution model to parametrize high-level policies that can invoke discrete options as well continuous control actions, and (2)  a cross-validation method that relaxes DDO's requirement that users specify the number of options to be discovered.
We evaluate \alg in simulation of a 3-link robot in the vertical plane pushing a block with friction and gravity, and in two physical experiments on the da Vinci surgical robot, needle insertion where a needle is grasped and inserted into a silicone tissue phantom, and needle bin picking where needles and pins are grasped from a pile and categorized into bins.
In the 3-link arm simulation, results suggest that \alg can take 3x fewer demonstrations to achieve the same reward compared to a baseline imitation learning approach.
In the needle insertion task, DDCO was successful 8/10 times compared to the next most accurate imitation learning baseline 6/10.
In the surgical bin picking task, the learned policy 
successfully grasps a single object in 66 out of 99 attempted grasps, and in all but one case successfully recovered from failed grasps by retrying a second time.
\end{abstract}



\section{Introduction}
An option is a short-term skill consisting of a control policy specialized for a specified region of the state space, and a termination condition recognizing leaving that region~\citep{suttonPS99}.
These short-term skills can be parametrized more concisely, e.g., by locally linear approximations to overall nonlinear motions, and these parameters can be substantially easier to learn, given the hierarchical structure.
The LfD and motion planning communities have long studied the problem of identifying and learning re-usable motion primitives for low-dimensional observation spaces~\citep{ijspreet2002learning,pastor2009learning,manschitz2015learning, buiVW02,krishnan2015transition,daniel2012hierarchical,krishnan2016swirl}.
A key question is how to extend these results to discover options that map high-dimensional observations to continuous controls and are parametrized by expressive deep neural networks.

Recently, several hierarchical reinforcement learning techniques have been proposed in deep reinforcement learning, which achieve state-of-the-art results on playing Atari games~\citep{baconHP16, fox2017ddo, vezhnevets2017feudal}.
One of these techniques is the Discovery of Deep Options (DDO) algorithm that takes a policy-gradient approach to train an Abstract Hidden Markov Model~\citep{buiVW02, danielVPN16} via Expectation-Gradient iterations.
This allows DDO to efficiently discover a specified number of \emph{deep options} from a set of demonstration trajectories, and train a policy for the task consisting of a high-level controller that selects from these options and the primitive actions~\citep{fox2017ddo}.

This paper presents an extension, called Discovery of Deep \emph{Continuous} Options (\alg), that extends the DDO algorithm to continuous control spaces and robotic imitation learning tasks. 
This requires two main algorithmic contributions.
First, we propose a parametrized nested representation which can be used to represent and train distributions over hybrid output spaces consisting of discrete options and continuous actions. 
Next, one limitation of DDO is that users must specify the number of options to be discovered.
Tuning this parameter online by exhaustive search can be infeasible in robotic settings which rely on physical evaluation.
We show that an efficient offline cross-validation step that maximizes the likelihood on a holdout set of demonstrations can automatically select the number of options. 
Empirically, we show that \alg learns options from demonstrations provided by human or algorithmic supervisors in simulation with a 3-link robot in the vertical plane pushing a block with friction and gravity, and in two physical experiments on the da Vinci surgical robot, needle insertion and needle bin picking.

\section{Related Work}
The concept of motion primitives, which are temporally-extended actions on a higher level of abstraction than direct motor control, is well studied~\citep{fikes1972learning}. 
Brooks described this architecture as ``layered control''~\citep{brooks1986robust}, and these ideas helped shape the field of hierarchical reinforcement learning~\citep{parr98,suttonPS99,barto03}.
The robotics community has recognized the need for closed-loop primitives, i.e., primitives that define control policies, and has proposed several control-theoretic models for defining and composing such primitives~\citep{ijspreet2002learning,pastor2009learning,manschitz2015learning, daniel2012hierarchical, lioutikov2015probabilistic}.
Similarly, task segmentation in Learning from Demonstrations considers partitioning a task based on demonstration data, which can be thought of as options~\citep{dayanH92,hengst02,kolterAN07,konidarisB07,niekum2012learning, calinon2014skills, konidarisKGB12, krishnan2016swirl, buiVW02, krishnan2015transition}.
However, all of these prior approaches consider low-dimensional observation spaces, and we look to extend these ideas to primitives that map high-dimensional observations, such as images, to controls.
There are also several approaches that consider ``demonstration clustering'', that is, identifying clusters of already segmented demonstrations~\citep{foxMT16, hausman2017multi, wang2017robust, wulfmeier2017mutual}.
We consider an algorithm that automatically segments and clusters the demonstrations.

Extending similar concepts to image data has been a recent area of interest, mostly in reinforcement and self-supervision settings for simulated domains~\citep{jonssonG16, vezhnevets2017feudal, florensaDA17, fox2017ddo}.
We explore leveraging these recent advances for imitation learning in realistic robotic manipulation tasks.
We extend one such algorithm from our prior work, DDO~\citep{fox2017ddo}, which uses a policy-gradient method to train an Abstract Hidden Markov Model~\citep{buiVW02, danielVPN16} via Expectation-Gradient iterations~\citep{salakhutdinov2003optimization,mclachlan2007algorithm}.
DDCO builds on DDO and on our prior work in task segmentation (Transition State Clustering~\citep{krishnan2015transition} and Sequential Windowed Inverse Reinforcement Learning~\citep{krishnan2016swirl}) with two new algorithmic extensions:  (1) a hybrid categorical--continuous distribution model to parametrize high-level policies that can invoke discrete options as well continuous control actions, and (2)  a cross-validation method that relaxes DDO's requirement that users specify the number of options to be discovered.

\section{Background: Imitation Learning and Options}
We consider the Imitation Learning (IL) setting, where a control policy is inferred from a dataset of demonstrations of the behavior of a human or algorithmic supervisor. We model the system as discrete-time, having at time $t$ a state $s_t$ in some state space $\C S$, such that applying the control $a_t$ in control space $\C A$ induces a stochastic state transition $s_{t+1}\sim p(s_{t+1}|s_t,a_t)$. The initial state is distributed $s_0\sim p_0(s_0)$. We assume that the set of demonstrations consists of trajectories, each a sequence of states and controls $\xi = (s_0, a_0, s_1, \ldots, s_T)$ of a given length $T$.

A flat, non-hierarchical policy $\pi_\theta(a_t \given s_t)$ defines the distribution over controls given the state, parametrized by $\theta\in\Theta$. In Behavior Cloning (BC), we train the parameter $\theta$ so that the policy fits the dataset of observed demonstrations and imitates the supervisor. For example, we can maximize the log-likelihood $L[\theta;\xi]$ that the stochastic process induced by the policy $\pi_\theta$ assigns to each demonstration trajectory $\xi$:
\eq{
L[\theta;\xi] = \log p_0(s_0) + \sum_{t=0}^{T-1} \log(\pi_\theta(a_t \given s_t)p(s_{t+1} \given s_t, a_t)).
}
When $\log\pi_\theta$ is parametrized by a deep neural network, we can perform stochastic gradient descent by sampling a batch of transitions, e.g. one complete trajectory, and computing the gradient
\eq{
\nabla_\theta L[\theta;\xi] = \sum_{t=0}^{T-1} \nabla_\theta \log \pi_\theta(a_t \given s_t).
}
Note that this method can be applied model-free, without any knowledge of the system dynamics $p$.
This paper considers $\pi_\theta(a_t|s_t)$ to be an isotropic Gaussian distribution, with a computed mean $\mu_\theta(s_t)$ and a fixed variance $\sigma^2$ (a hyper-parameter that can be set by cross-validation), simplifying the gradient to the standard quadratic loss:
\eqn{\label{eq:square}
\nabla_\theta L[\theta;\xi] = -\sum_{t=0}^{T-1} \nabla_\theta \frac{(\mu_\theta(s_t)-a_t)^2}{2\sigma^2} = \sum_{t=0}^{T-1} \left(\frac{a_t - \mu_\theta(s_t)}{\sigma^2}\right)^\T \nabla_\theta \mu_\theta(s_t).
}

In the options framework~\citep{suttonPS99}, an option represents a low-level policy that can be invoked by a high-level policy to perform a certain sub-task. Formally, an option $h$ in an options set $\C H$ is specified by a control policy $\pi_h(a_t \given s_t)$ and a stochastic termination condition $\psi_h(s_t)\in[0,1]$. The high-level policy $\eta(h_t|s_t)$ defines the distribution over options given the state.
Once an option $h$ is invoked, physical controls are selected by the option's policy $\pi_h$ until it terminates. After each physical control is applied and the next state $s'$ is reached, the option $h$ terminates with probability $\psi_h(s')$, and if it does then the high-level policy selects a new option $h'$ with distribution $\eta(h'|s')$.
Thus the interaction of the hierarchical control policy $\lan\eta,(\pi_h,\psi_h)_{h\in\C H}\ran$ with the system induces a stochastic process over the states $s_t$, the options $h_t$, the controls $a_t$, and the binary termination indicators $b_t$.


\section{Discovery of Deep Continuous Options (\alg)}
In this section, we give an overview of the recently proposed Discovery of Deep Options (DDO) algorithm~\citep{fox2017ddo}, and present its extension to continuous control spaces, as well as a cross-validation method for selecting the number of options to be discovered.

\subsection{Discovery of Deep Options (DDO)}
DDO is a policy-gradient algorithm that discovers deep options by fitting their parameters to maximize the likelihood of a set of demonstration trajectories.
We denote by $\theta$ the vector of all trainable parameters used for $\eta$ and for $\pi_h$ and $\psi_h$ of each option $h\in\C H$. 
We wish to find the $\theta\in\Theta$ that maximizes the log-likelihood of generating each demonstration trajectory $\xi=(s_0,a_0,s_1,\ldots,s_T)$. The challenge is that this log-likelihood depends on the latent variables in the stochastic process, namely the options and the termination indicators $\zeta = (b_0,h_0,b_1,h_1,\ldots,h_{T-1})$. 
DDO employs the Expectation Gradient trick~\citep{salakhutdinov2003optimization,mclachlan2007algorithm}:
\eq{
\nabla_\theta L[\theta;\xi] = \E_\theta[\nabla_\theta \log \p_\theta(\zeta,\xi) \given \xi],
}
where $\p_\theta(\zeta,\xi)$ is the joint probability of the latent and observable variables, given by
\eq{
\p_\theta(\zeta,\xi) = p_0(s_0) \delta_{b_0=1}\eta(h_0 \given s_0) \prod_{t=1}^{T-1} \p_\theta(b_t, h_t \given h_{t-1}, s_t) \prod_{t=0}^{T-1} \pi_{h_t}(a_t \given s_t) p(s_{t+1} \given s_t, a_t) ,
}
where in the latent transition $\p_\theta(b_t, h_t \given h_{t-1}, s_t)$ we have with probability $\psi_{h_{t-1}}(s_t)$ that $b_t=1$ and $h_t$ is drawn from $\eta(\cdot|s_t)$, and otherwise that $b_t=0$ and $h_t$ is unchanged, i.e.
\eq{
\p_\theta(b_t {=} 1, h_t \given h_{t-1}, s_t) &= \psi_{h_{t-1}}(s_t) \eta(h_t \given s_t) \\
\p_\theta(b_t {=} 0, h_t \given h_{t-1}, s_t) &= (1 - \psi_{h_{t-1}}(s_t)) \delta_{h_t = h_{t-1}}.
}
Expectation-Gradient follows an alternating optimization scheme similar to Expectation-Maximization (EM), with an E-step and a G-step. Instead of a maximization step as in EM, the EG algorithm takes a gradient step with respect to the parameters. 
%
The E-step computes the marginal posteriors
\eq{
u_t(h) = \p_\theta(h_t {=} h \given \xi); && v_t(h) = \p_\theta(b_t {=} 1, h_t {=} h \given \xi); && w_t(h) = \p_\theta(h_t {=} h, b_{t+1} {=} 0 \given \xi)
}
using a forward-backward algorithm similar to Baum-Welch~\citep{baum1972equality}, and the G-step uses these posteriors to compute the log-likelihood gradient

\eqn{\label{eq:gradient}
\nabla_\theta L[\theta;\xi] = \sum_{h\in\C H} \Biggl(& \sum_{t=0}^{T-1}\mrlap{ \Biggl(v_t(h) \nabla_\theta \log \eta(h \given s_t) +  u_t(h)\nabla_\theta \log \pi_h(a_t \given s_t)\Biggr)} \\ 
& + \sum_{t=0}^{T-2} \Biggl((u_t(h)-w_t(h)) \nabla_\theta \log \psi_h(s_{t+1}) + w_t(h) \nabla_\theta \log (1 - \psi_h(s_{t+1})) \Biggr)\Biggr).\nn
}
The DDO algorithm is derived in detail in~\citep{fox2017ddo}.
The gradient computed above can then be used in any stochastic gradient descent algorithm. In our experiments we use Adam and Momentum. 

In its original formulation, DDO is applied to finite action spaces, and $\log\pi_h$ is the log-probability of a categorical distribution. We note that DDO can be applied to continuous control spaces, by instead computing the log-density of the continuous distribution $\pi_h(a_t|s_t)$, which in the case of isotropic Gaussian distributions is proportional to the square distance of $a_t$ from the mean control $\mu_h(s_t)$, as in~\eqref{eq:square}.

\subsection{Hybrid Categorical--Continuous Distribution Model}
\slab{hybrid}
We allow the high-level policy to solve some sub-tasks directly, without invoking an option, by augmenting its output space to include both physical control and options, $\C A\cup\C H$. In states that are not clustered into any option, the high-level policy can choose the physical control to apply in every step.
In continuous control spaces, this introduces the challenge that $\eta(h_t|s_t)$ is now a hybrid categorical--continuous distribution. We denote by $\eta(h^c|s)=1-\sum_{h\in\C H}\eta(h|s)$ the probability that the high level applies physical control, and by $\mu_\eta(s)$ the mean control in that event. Then the continuous part of the distribution has the unnormalized density $\eta(a|s)=\eta(h^c|s)\C N(a;\mu_\eta(s),\sigma^2)$.



This allows us to represent and train the hybrid distribution with the following neural network architecture. With $k=|\C H|$, the output layer represents a categorical distribution with log-probabilities $y_0,\ldots,y_k$, as well as the parameter of the control distribution $\mu_\eta$. The relevant gradient terms in~\eqref{eq:gradient} are, as before, $v_t(h)\nabla_\theta y_h(s_t$, and now additionally the complement term $v_t(h^c)\nabla y_0(s_t)$, with the probability $v_t(h^c)=\p_\theta(b_t{=}1)-\sum_{h\in\C H}v_t(h)$ that the high level applies physical control in time-step $t$; as well as the control-gradient term $v_t(h^c)\left(\frac{a_t-\mu_\eta(s_t)}{\sigma^2}\right)^\T\nabla_\theta\mu_\eta(s_t)$, similarly to~\eqref{eq:square}.

\subsection{Cross-Validation For Parameter Tuning}
\slab{k-validation}
We explored whether it is possible to tune the number of options offline.
\alg is based on a maximum-likelihood formulation, which describes the likelihood that the observed demonstrations are generated by a hierarchy parametrized by $\theta$.
However, the model expressiveness is strictly increasing in $k$, causing the optimal training likelihood to increase even beyond the point where the model overfits to the demonstrations and fails to generalize to unseen states.
We therefore adopt a cross-validation technique that holds out 10\% of the demonstration trajectories for each of $10$ folds, trains on the remaining data, and validates the trained model on the held out data.
We select the value of $k$ that achieves the highest average log-likelihood over the $10$ folds, suggesting that training such a hierarchical model generalizes well.
We train the final policy over the entire data.
We find empirically that the cross-validated log-likelihood serves as a good proxy to actual task performance.

\section{Results}

\subsection{Box2D Simulation: 2D Surface Pushing with Friction and Gravity}
\begin{figure}[t]\vspace{-2em}
    \centering
    \includegraphics[width=\textwidth]{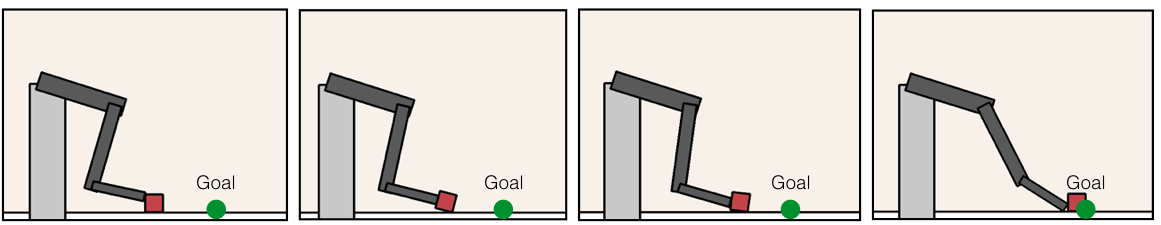}
    \caption{A 3-link robot arm has to push a box along a surface with friction to a randomly chosen goal state to the box's left or right without toppling it. Due to the collision and contact constraints of this task, it has two geometrically distinct pushing skills, backhand and forehand, for goals on the left and on the right. \label{fig:b2dexp}} 
\end{figure}

In the first experiment, we simulate a 3-link robot arm in Box2D (Figure \ref{fig:b2dexp}). This arm consists of three links of lengths 5 units, 5 units, and 3 units, connected by ideal revolute joints. The arm is controlled by setting the values of the joint angular velocities $\dot{\phi}_1, \dot{\phi}_2, \dot{\phi}_3$. In the environment, there is a box that lies on a flat surface with uniform friction. The objective is to push this box without toppling it until it rests in a randomly chosen goal position. After this goal state is reached, the goal position is regenerated randomly.
The task is for the robot to push the box to as many goals as possible in 2000 time-steps.
Our algorithmic supervisor runs the RRT Connect motion planner of the Open Motion Planning Library, \textsf{ompl}, at each time-step planning to reach the goal. 
Due to the geometry of the configuration space and the task, there are two classes of trajectories that are generated, when the goal is to the left or right of the arm. 
Details of the environment and motion planner are included in the supplementary material (\textbf{SM 1.1}).

\begin{figure}[t]
    \centering
    \includegraphics[width=\textwidth]{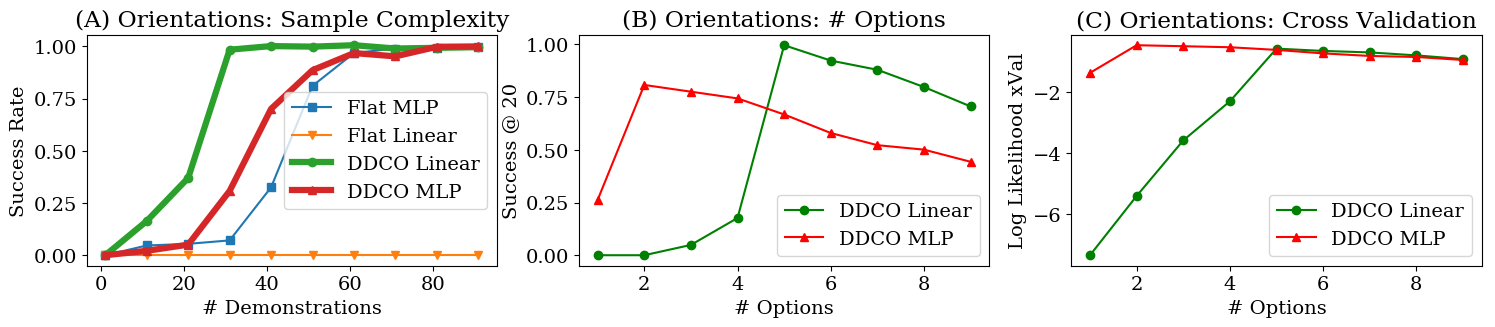}
    \includegraphics[width=\textwidth]{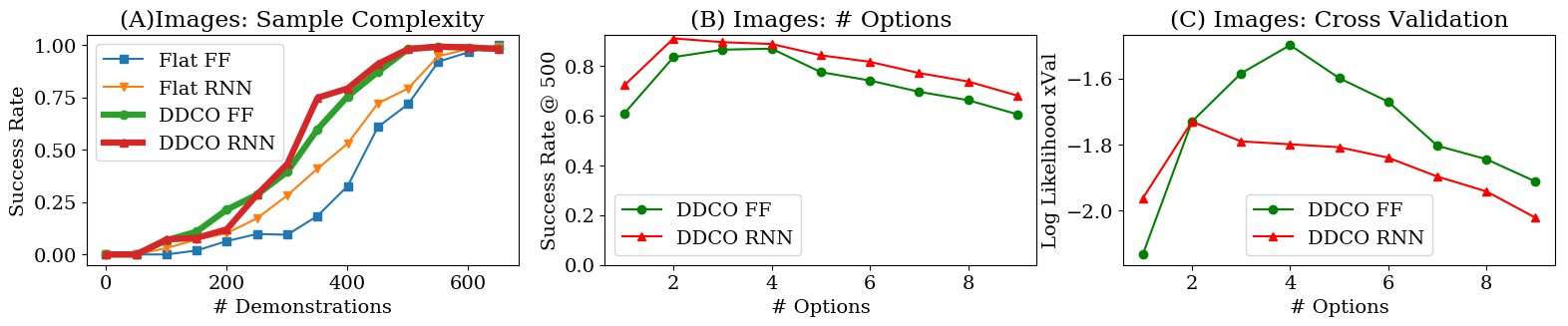}
    \caption{2D Surface Pushing from low-dimensional observations and image observations. (A) The sample complexity of different policy representations, (B) The success rate for different numbers of discovered options, (C) The peak in the reward correlates over the number of options with the 10-fold cross-validated log-likelihood. \label{fig:b2dexp1-1}}
\end{figure}

\vspace{0.25em}\noindent\textbf{Observing Positions and Orientations: } First, we consider the case where we observe the full low-dimensional state of the system: three joint angles, the box's position and orientation, and the position of the goal state. 
We compare our hierarchical policies with two flat, non-hierarchical policies. One baseline policy we consider is an under-parametrized linear policy which is not expressive enough to approximate the supervisor policy well. The other baseline policy is a multi-layer perceptron (MLP) policy.
We train each of these policies via Behavior Cloning (BC), i.e. by maximizing the likelihood each gives to the set of demonstrations.
As expected, the flat linear policy is unsuccessful at the task for any number of observed demonstrations (Figure \ref{fig:b2dexp1-1} Top-A). The MLP policy, on the other hand, can achieve the maximum reward when trained on 60 demonstrations or more. 

We apply \alg and learn two 2-level hierarchical policies, one with linear low-level options, and the other with MLP low-level options of the same architecture used for the flat policy. 
In both cases, the termination conditions are parametrized by a logistic regression from the state to the termination probability, and the high-level policy is a logistic regression from the state to an option selection.
For the linear hierarchy we set \alg to discover 5 options, and for the MLP hierarchy we discover 2 options.
The MLP hierarchical policy can achieve the maximum reward with  30 demonstrations, and is therefore 2x more sample-efficient than its flat counterpart (Figure \ref{fig:b2dexp1-1} Top A).
Details of the parametrization are included in the Supplementary Material (\textbf{SM 1.1}).

We also vary the number of options discovered by \alg, and plot the reward obtained by the resulting policy (Figure \ref{fig:b2dexp1-1} Top-B). While the performance is certainly sensitive to the number of options, we find that the benefit of having sufficiently many options is only diminished gradually with each additional option beyond the optimum. Importantly, the peak in the cross-validated log-likelihood corresponds to the number of options that achieves the maximum reward (Figure \ref{fig:b2dexp1-1} Top-C). This allows us to use cross-validation to select the number of options without having to evaluate the policy by rolling it out in the environment.

\vspace{0.25em}\noindent\textbf{Observing the Image: } Next, we consider the case where the sensor inputs are 640x480 images of the scene. The low-dimensional state is still fully observable in these images, however these features are not observed explicitly, and must be extracted by the control policy. We consider two neural network architectures to represent the policy: a convolutional layer followed by either a fully connected layer or an LSTM, respectively forming a feed-forward (FF) network and a recurrent network (RNN).
We use these architectures both for flat policies and for low-level options in hierarchical policies.
For the hierarchical policies, as in the previous experiment, we consider a high-level policy that only selects options.
For the FF policy we discover 4 options, and for the RNN policy we discover 2 options.
The details of the parametrization are included in the Supplementary Material (\textbf{SM 1.2}).

\begin{figure}[t]\vspace{-2em}
    \centering
    \includegraphics[width=0.8\textwidth]{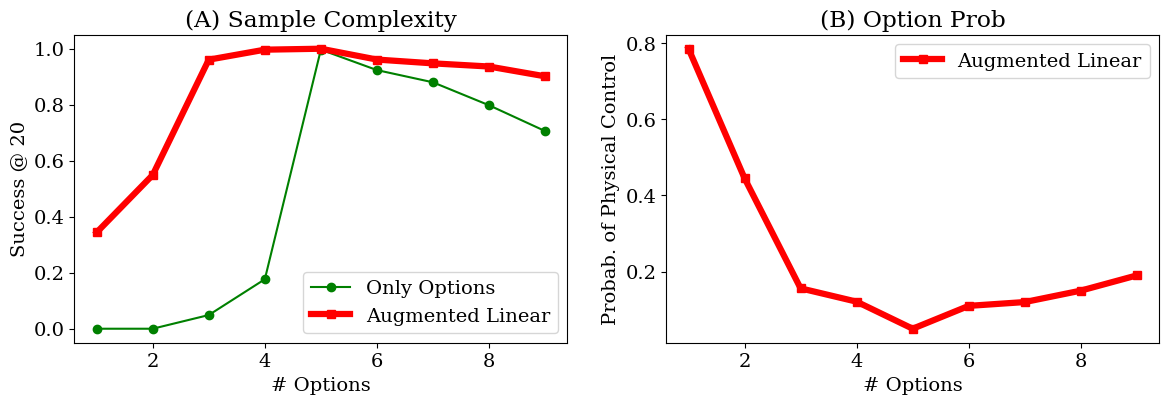}
    \caption{ (A) The sample complexity of augmenting the high-level control space v.s. selecting only options. (B) The fraction of high-level selections that are of physical control. As the number of options increases the frequency of selecting a physical control goes down. \label{fig:b2dexp1-3}}
\end{figure}

The hierarchical policies require 30\% fewer demonstrations than the flat policies to achieve the maximum reward (Figure \ref{fig:b2dexp1-1} Bottom A).
Figure \ref{fig:b2dexp1-1} Bottom-B and Figure \ref{fig:b2dexp1-1} Bottom-C show how the success rate and cross-validated log-likelihood vary with the number of options.
As for the low-dimensional inputs, the success rate curve is correlated with the cross-validated log-likelihood.
We can rely on this to select the number of options offline without rolling out the learned hierarchical policy.

\vspace{0.25em}\noindent\textbf{Control Space Augmentation: } 
We test two different architectures for the output layer of the high-level policy: either a softmax categorical distribution selecting an option, or the hybrid categorial--continuous distribution output described in~\sref{hybrid}.
The low-level policies are linear.

Figure \ref{fig:b2dexp1-3} describes the empirical estimation, through policy rollouts, of the success rate as a function of the number of options, and of the fraction of time that the high-level policy applies physical control.
When the options are too few to provide skills that are useful throughout the state space, physical controls can be selected instead to compensate in states where no option would perform well. This is indicated by physical control being selected with greater frequency. As more options allow a better coverage of the state space, the high-level policy selects physical controls less often, allowing it to generalize better by focusing on correct option selection. With too many discovered options, each option is trained from less data on average, making some options overfit and become less useful to the high-level policy.
In the Supplementary Material (\textbf{SM 1.5}), we perform the same experiment for the image-based case, with similar results.

\subsection{Physical Experiment 1: Needle Insertion}
\begin{figure}
    \centering
    \includegraphics[width=0.8\textwidth]{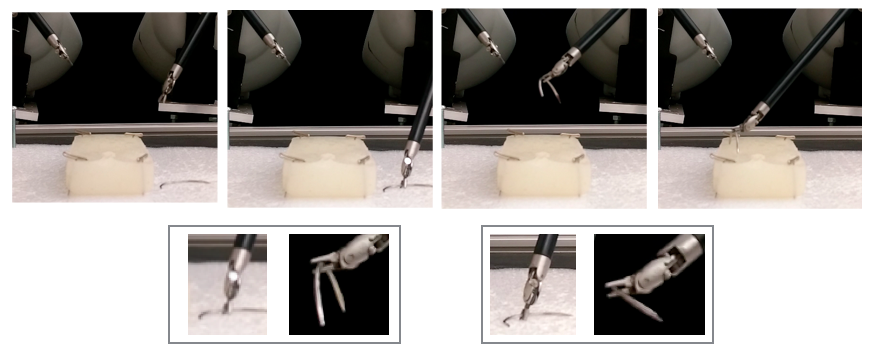}
    \caption{A needle orienting and insertion task. The robot must grasp a surgical needle, reorient it in parallel to a tissue phantom, and insert the needle.   \label{fig:dvrkexp3}}
\end{figure}

We consider a needle orientation and insertion task on the da Vinci Research Kit (DVRK) surgical robot (Figure \ref{fig:dvrkexp3}). In this task, the robot must grasp a surgical needle, reorient it in parallel to a tissue phantom, and insert the needle into the tissue phantom.
The task is successful if the needle is inserted into a 1cm diameter target region on the phantom. Small changes in the needle's initial orientation can lead to large changes to the in-gripper pose of the needle due to deflection. 
The state is the current 6-DoF pose of the robot gripper, and algorithmically extracted visual features that describe the estimated pose of the needle.
These features are derived from an image segmentation that masks the needle from the background and fits an ellipsoid to the resulting pixels. 
The principal axis of this 2D ellipsoid is a proxy for the pose of the needle.
The task runs for a fixed 15 time-steps, and the policy must set the joint angles of the robot at each time-step.

The needle's deflection coupled with the inaccurate kinematics of the DVRK make it challenging to plan trajectories to insert the needle properly.
A visual servoing policy needs to be trained that can both grasp the needle in the correct position, as well as reorient the gripper in the correct direction after grasping.
To collect demonstrations, we programmed an initial open-loop control policy, interrupted the robot via keyboard input when adjustment was needed, and kinesthetically adjusted the pose of the gripper. 
We collected 100 such demonstrations.

 We evaluated the following alternatives: (1) a single flat MLP policy with continuous output, (2) a flat policy consisting of 15 distinct MLP networks, one for each time-step, (3) a hierarchical policy with 5 options trained with \alg. We considered a hierarchy where the high-level policy is a MLP with a softmax output that selects the appropriate option, and each option is parametrized by a distinct MLP with continuous outputs.
 \alg learns 5 options, two of which roughly correspond to the visual servoing for grasping and lifting the needle, and the other three handle three different types of reorientation.
 For 100 demonstrations, the hierarchical policy learned with \alg has a 45\%  higher log-likelihood measured in cross-validation than the flat policy, and a 24\% higher log-likelihood than the per-timestep policy. 
  Training results, hyper-parameter tuning, interpretation of the options learned, and details of the parametrization are included in the Supplementary Material (\textbf{SM 1.3}).

We ran preliminary trials to confirm that the trained options can be executed on the robot.
For each of the methods, we report the success rate in 10 trials, i.e. the fraction of trials in which the needle was successfully grasped and inserted in the target region. 
All of the techniques had comparable accuracy in trials where they successfully grasped and inserted the needle into the 1cm diameter target region.
The algorithmic open-loop policy only succeeded 2/10 times.
Surprisingly, Behavior Cloning (BC) did not do much better than the open-loop policy, succeeding only 3/10 times.
Per-timestep BC was far more successful (6/10).
Finally, the hierarchical policy learned with \alg succeeded 8/10 times. On 10 trials it was successful 5 times more than the direct BC approach and 2 times more than the per-timestep BC approach.
While not statistically significant, our preliminary results suggest that hierarchical imitation learning is also beneficial in terms of task success, in addition to improving model generalization and interpretability.
Detailed discussion of the particular failure modes is included in the Supplementary Material (\textbf{SM 1.3}).

\subsection{Physical Experiment 2: Surgical Bin Picking}
In this task, the robot is given a foam bin with a pile of 5--8 needles of three different types, each 1--3mm in diameter.
The robot must extract needles of a specified type and place them in an ``accept'' cup, while placing all other needles in a ``reject'' cup.
The task is successful if the entire foam bin is cleared into the correct cups.

In initial trials, the kinematics of the DVRK were not precise enough for grasping needles.
We then realized that visual servoing is needed, which requires learning.
However, even with visual servoing, failures are common, and we would like to also learn automatic recovery behaviors. 
To define the state space for this task, we first generate binary images from overhead stereo images, and apply a color-based segmentation to identify the needles (the \textsf{image} input).
Then, we use a classifier trained in advance on 40 hand-labeled images to identify and provide a candidate grasp point, specified by position and direction in image space (the \textsf{grasp} input). 
Additionally, the 6 DoF robot gripper pose and the open-closed state of the gripper are observed (the \textsf{kin} input).
The state space of the robot is (\textsf{image}, \textsf{grasp}, \textsf{kin}), and the control space is the 6 joint angles and the gripper angle.

Each sequence of grasp, lift, move, and drop operations is implemented in 10 control steps of joint angle positions.
As in the previous task,  we programmed an initial open-loop control policy, interrupted the robot via keyboard input when adjustment was needed, and kinesthetically adjusted the pose of the gripper. 
We collected 60 such demonstrations, in each fully clearing a pile of 3--8 needles from the bin, for a total of 450 individual grasps.
Details of the state space and the demonstration protocol are included in the Supplementary Material (\textbf{SM 1.4}).

\begin{figure}\vspace{-2em}
    \includegraphics[width=0.58\textwidth]{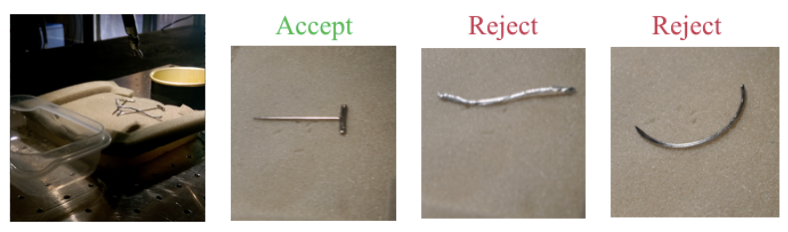}
    \includegraphics[width=0.4\textwidth]{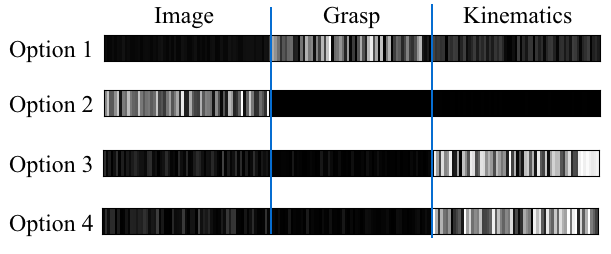}
    \caption{(Left) Surgical Bin Picking: the robot is given a foam bin with a pile of 5--8 needles of three different types, each 1--3mm in diameter. The robot must extract needles of a specified type and place them in an ``accept'' cup, while placing all other needles in a ``reject'' cup. (Right) For each of the 4 options, we plot how heavily the different inputs are weighted (\textsf{image}, \textsf{grasp}, or \textsf{kin}) in computing the option's action. Nonzero values of the ReLU units are marked in white and indicate input relevance. \label{fig:dvrkexp4}} 
\end{figure}

We apply \alg to the collected demonstrations with the policy architecture described in \textbf{SM 1.4}. In this network, there are three inputs: a binary \textsf{image}, a candidate \textsf{grasp}, and kinematics (\textsf{kin}). These inputs are processed by distinct branches of the network before being aggregated into a single feature layer.  Using the cross-validation technique described in~\sref{k-validation}, we selected the number of options to be $k=4$.

We plot the average activations of the feature layer for each low-level option (Figure \ref{fig:dvrkexp4}). While this is only a coarse analysis, it gives some indication of which inputs (\textsf{image}, \textsf{grasp}, or \textsf{kin}) are relevant to the policy and termination. We see that the options are clearly specialized. The first option has a strong dependence only on the \textsf{grasp} candidate, the second option attends almost exclusively to the \textsf{image}, while the last two options rely mostly on \textsf{kin}.
 Details of the architecture and additional interpretations of the learned options are included in the Supplementary Material (\textbf{SM 1.4}).

Finally, we evaluate the success of the learned hierarchical policy in 10 full trials, according to 4 different success criteria. First, the overall task success is measured by the success rate of fully clearing the bin without failing 4 consecutive grasping attempts. Second, we measure the success rate of picking exactly one needle in individual grasp attempts. Third, we measure the success rate of appropriately reacting to grasps, by dropping the load and retrying unsuccessful grasps, and not dropping successful grasps. Fourth, we measured the success rate of needle categorization.

In 10 trials, 7/10 were successful. The main failure mode was unsuccessful grasping due to picking either no needles or multiple needles. As the piles were cleared and became sparser, the robot's grasping policy became somewhat brittle. The grasp success rate was 66\% on 99 attempted grasps. In contrast, we rarely observed failures at the other aspects of the task, reaching 97\% successful recovery on 34 failed grasps. The Supplementary Material (\textbf{SM 1.4}) describes a full breakdown of the failure modes, as well as an additional experiment of policy generalization to unseen objects.

\section{Conclusion}
\label{sec:conclusion}
This paper extends our prior work on the DDO algorithm, which was designed for discrete action spaces, to the continuous robot control problems, and proposes new methods for addressing issues of hybrid discrete-continuous action parametrization and hyper-parameter tuning for the number of options.
Results suggest that imposing a hierarchical structure with DDCO on the trained policy has the potential to significantly reduce the number of demonstrations needed for learning in real and simulated tasks.
DDCO facilitates imitation learning for long duration, multi-step tasks such as bin picking which involves grasping, categorizing, and recovery behaviors.
We believe that this hierarchical structure can also facilitate generalization in multi-task learning settings, and hope to explore that in further detail in future work.
In future work, we also hope to explore heterogeneous option parametrizations, where some options are learned and some are derived from analytic formulae.



\clearpage
\acknowledgments{This research was performed at the AUTOLAB at UC Berkeley and the Real-Time Intelligent Secure Execution (RISE) Lab in affiliation with the Berkeley AI Research (BAIR) Lab and the CITRIS "People and
Robots" (CPAR) Initiative.
This research is supported in part by DHS Award HSHQDC-16-3-00083, NSF CISE Expeditions Award CCF-1139158, by the Scalable Collaborative Human-Robot Learning (SCHooL) Project NSF National Robotics Initiative Award 1734633, and donations from Alibaba, Amazon, Ant Financial, CapitalOne, Ericsson, GE, Google, Huawei, Intel, IBM, Microsoft, Scotiabank, VMware, Siemens, Cisco, Autodesk, Toyota Research, Samsung, Knapp, and Loccioni Inc. We also acknowledge a major equipment grant from Intuitive Surgical and by generous donations from Andy Chou and Susan and Deepak Lim. We thank our colleagues who provided helpful feedback and suggestions, in particular Jeff Mahler and Michael Laskey.}


\bibliography{robotics,ml}  

\begin{thebibliography}{33}
\providecommand{\natexlab}[1]{#1}
\providecommand{\url}[1]{\texttt{#1}}
\expandafter\ifx\csname urlstyle\endcsname\relax
  \providecommand{\doi}[1]{doi: #1}\else
  \providecommand{\doi}{doi: \begingroup \urlstyle{rm}\Url}\fi

\bibitem[Sutton et~al.(1999)Sutton, Precup, and Singh]{suttonPS99}
R.~S. Sutton, D.~Precup, and S.~P. Singh.
\newblock Between {MDP}s and semi-{MDP}s: {A} framework for temporal
  abstraction in reinforcement learning.
\newblock \emph{{AI}}, 112\penalty0 (1-2):\penalty0 181--211, 1999.
\newblock \doi{10.1016/S0004-3702(99)00052-1}.

\bibitem[Ijspeert et~al.(2002)Ijspeert, Nakanishi, and
  Schaal]{ijspreet2002learning}
A.~Ijspeert, J.~Nakanishi, and S.~Schaal.
\newblock Learning attractor landscapes for learning motor primitives.
\newblock In \emph{Neural Information Processing Systems (NIPS)}, pages
  1523--1530, 2002.

\bibitem[Pastor et~al.(2009)Pastor, Hoffmann, Asfour, and
  Schaal]{pastor2009learning}
P.~Pastor, H.~Hoffmann, T.~Asfour, and S.~Schaal.
\newblock Learning and generalization of motor skills by learning from
  demonstration.
\newblock In \emph{IEEE ICRA}, 2009.

\bibitem[Manschitz et~al.(2015)Manschitz, Kober, Gienger, and
  Peters]{manschitz2015learning}
S.~Manschitz, J.~Kober, M.~Gienger, and J.~Peters.
\newblock Learning movement primitive attractor goals and sequential skills
  from kinesthetic demonstrations.
\newblock \emph{Robotics and Autonomous Systems}, 2015.

\bibitem[Bui et~al.(2002)Bui, Venkatesh, and West]{buiVW02}
H.~H. Bui, S.~Venkatesh, and G.~West.
\newblock Policy recognition in the abstract hidden {M}arkov model.
\newblock \emph{{JAIR}}, 17:\penalty0 451--499, 2002.

\bibitem[Krishnan et~al.(2015)Krishnan, Garg, Patil, Lea, Hager, Abbeel, and
  Goldberg]{krishnan2015transition}
S.~Krishnan, A.~Garg, S.~Patil, C.~Lea, G.~Hager, P.~Abbeel, and K.~Goldberg.
\newblock Transition state clustering: Unsupervised surgical trajectory
  segmentation for robot learning.
\newblock In \emph{International Symposium of Robotics Research. Springer
  STAR}, 2015.

\bibitem[Daniel et~al.(2012)Daniel, Neumann, and
  Peters]{daniel2012hierarchical}
C.~Daniel, G.~Neumann, and J.~Peters.
\newblock Hierarchical relative entropy policy search.
\newblock In \emph{AISTATS}, pages 273--281, 2012.

\bibitem[Krishnan et~al.()Krishnan, Garg, Liaw, Thananjeyan, Miller, Pokorny,
  and Goldberg]{krishnan2016swirl}
S.~Krishnan, A.~Garg, R.~Liaw, B.~Thananjeyan, L.~Miller, F.~T. Pokorny, and
  K.~Goldberg.
\newblock Swirl: A sequential windowed inverse reinforcement learning algorithm
  for robot tasks with delayed rewards.

\bibitem[Bacon et~al.(2016)Bacon, Harb, and Precup]{baconHP16}
P.-L. Bacon, J.~Harb, and D.~Precup.
\newblock The option-critic architecture.
\newblock \emph{arXiv preprint arXiv:1609.05140}, 2016.

\bibitem[Fox et~al.(2017)Fox, Krishnan, Stoica, and Goldberg]{fox2017ddo}
R.~Fox, S.~Krishnan, I.~Stoica, and K.~Goldberg.
\newblock Multi-level discovery of deep options.
\newblock \emph{arXiv preprint arXiv:1703.08294}, 2017.

\bibitem[Vezhnevets et~al.(2017)Vezhnevets, Osindero, Schaul, Heess, Jaderberg,
  Silver, and Kavukcuoglu]{vezhnevets2017feudal}
A.~S. Vezhnevets, S.~Osindero, T.~Schaul, N.~Heess, M.~Jaderberg, D.~Silver,
  and K.~Kavukcuoglu.
\newblock Feudal networks for hierarchical reinforcement learning.
\newblock \emph{arXiv preprint arXiv:1703.01161}, 2017.

\bibitem[Daniel et~al.(2016)Daniel, Van~Hoof, Peters, and Neumann]{danielVPN16}
C.~Daniel, H.~Van~Hoof, J.~Peters, and G.~Neumann.
\newblock Probabilistic inference for determining options in reinforcement
  learning.
\newblock \emph{Machine Learning}, 104\penalty0 (2-3):\penalty0 337--357, 2016.

\bibitem[Fikes et~al.(1972)Fikes, Hart, and Nilsson]{fikes1972learning}
R.~E. Fikes, P.~E. Hart, and N.~J. Nilsson.
\newblock Learning and executing generalized robot plans.
\newblock \emph{Artificial intelligence}, 3:\penalty0 251--288, 1972.

\bibitem[Brooks(1986)]{brooks1986robust}
R.~Brooks.
\newblock A robust layered control system for a mobile robot.
\newblock \emph{IEEE journal on robotics and automation}, 2\penalty0
  (1):\penalty0 14--23, 1986.

\bibitem[Parr(1998)]{parr98}
R.~E. Parr.
\newblock \emph{Hierarchical control and learning for Markov decision
  processes}.
\newblock PhD thesis, UNIVERSITY of CALIFORNIA at BERKELEY, 1998.

\bibitem[Barto and Mahadevan(2003)]{barto03}
A.~G. Barto and S.~Mahadevan.
\newblock Recent advances in hierarchical reinforcement learning.
\newblock \emph{Discrete Event Dynamic Systems}, 13\penalty0 (1-2):\penalty0
  41--77, 2003.
\newblock \doi{10.1023/A:1022140919877}.

\bibitem[Lioutikov et~al.(2015)Lioutikov, Neumann, Maeda, and
  Peters]{lioutikov2015probabilistic}
R.~Lioutikov, G.~Neumann, G.~Maeda, and J.~Peters.
\newblock Probabilistic segmentation applied to an assembly task.
\newblock In \emph{Humanoid Robots (Humanoids), 2015 IEEE-RAS 15th
  International Conference on}, pages 533--540. IEEE, 2015.

\bibitem[Dayan and Hinton(1992)]{dayanH92}
P.~Dayan and G.~E. Hinton.
\newblock Feudal reinforcement learning.
\newblock In \emph{{NIPS}}, pages 271--278, 1992.

\bibitem[Hengst(2002)]{hengst02}
B.~Hengst.
\newblock Discovering hierarchy in reinforcement learning with {HEXQ}.
\newblock In \emph{{ICML}}, volume~2, pages 243--250, 2002.

\bibitem[Kolter et~al.(2007)Kolter, Abbeel, and Ng]{kolterAN07}
J.~Z. Kolter, P.~Abbeel, and A.~Y. Ng.
\newblock Hierarchical apprenticeship learning with application to quadruped
  locomotion.
\newblock In \emph{NIPS}, volume~20, 2007.

\bibitem[Konidaris and Barto(2007)]{konidarisB07}
G.~Konidaris and A.~G. Barto.
\newblock Building portable options: Skill transfer in reinforcement learning.
\newblock In \emph{IJCAI}, volume~7, pages 895--900, 2007.

\bibitem[Niekum et~al.(2012)Niekum, Osentoski, Konidaris, and
  Barto]{niekum2012learning}
S.~Niekum, S.~Osentoski, G.~Konidaris, and A.~Barto.
\newblock Learning and generalization of complex tasks from unstructured
  demonstrations.
\newblock In \emph{Int. Conf. on Intelligent Robots and Systems (IROS)}. IEEE,
  2012.

\bibitem[Calinon(2014)]{calinon2014skills}
S.~Calinon.
\newblock Skills learning in robots by interaction with users and environment.
\newblock In \emph{IEEE Int. Conf. on Ubiquitous Robots and Ambient
  Intelligence (URAI)}, 2014.

\bibitem[Konidaris et~al.(2012)Konidaris, Kuindersma, Grupen, and
  Barto]{konidarisKGB12}
G.~Konidaris, S.~Kuindersma, R.~A. Grupen, and A.~G. Barto.
\newblock Robot learning from demonstration by constructing skill trees.
\newblock \emph{{IJRR}}, 31\penalty0 (3):\penalty0 360--375, 2012.
\newblock \doi{10.1177/0278364911428653}.

\bibitem[Fox et~al.(2016)Fox, Moshkovitz, and Tishby]{foxMT16}
R.~Fox, M.~Moshkovitz, and N.~Tishby.
\newblock Principled option learning in {M}arkov decision processes.
\newblock In \emph{UAI}, 2016.

\bibitem[Hausman et~al.(2017)Hausman, Chebotar, Schaal, Sukhatme, and
  Lim]{hausman2017multi}
K.~Hausman, Y.~Chebotar, S.~Schaal, G.~Sukhatme, and J.~Lim.
\newblock Multi-modal imitation learning from unstructured demonstrations using
  generative adversarial nets.
\newblock \emph{arXiv preprint arXiv:1705.10479}, 2017.

\bibitem[Wang et~al.(2017)Wang, Merel, Reed, Wayne, de~Freitas, and
  Heess]{wang2017robust}
Z.~Wang, J.~Merel, S.~Reed, G.~Wayne, N.~de~Freitas, and N.~Heess.
\newblock Robust imitation of diverse behaviors.
\newblock \emph{arXiv preprint arXiv:1707.02747}, 2017.

\bibitem[Wulfmeier et~al.(2017)Wulfmeier, Posner, and
  Abbeel]{wulfmeier2017mutual}
M.~Wulfmeier, I.~Posner, and P.~Abbeel.
\newblock Mutual alignment transfer learning.
\newblock \emph{arXiv preprint arXiv:1707.07907}, 2017.

\bibitem[Jonsson and G{\'o}mez(2016)]{jonssonG16}
A.~Jonsson and V.~G{\'o}mez.
\newblock Hierarchical linearly-solvable {M}arkov decision problems.
\newblock \emph{arXiv preprint arXiv:1603.03267}, 2016.

\bibitem[Florensa et~al.(2017)Florensa, Duan, and Abbeel]{florensaDA17}
C.~Florensa, Y.~Duan, and P.~Abbeel.
\newblock Cstochastic neural networks for hierarchical reinforcement learning.
\newblock In \emph{{ICLR}}, 2017.

\bibitem[Salakhutdinov et~al.(2003)Salakhutdinov, Roweis, and
  Ghahramani]{salakhutdinov2003optimization}
R.~Salakhutdinov, S.~Roweis, and Z.~Ghahramani.
\newblock Optimization with {EM} and expectation-conjugate-gradient.
\newblock In \emph{{ICML}}, pages 672--679, 2003.

\bibitem[McLachlan and Krishnan(2007)]{mclachlan2007algorithm}
G.~McLachlan and T.~Krishnan.
\newblock \emph{The {EM} algorithm and extensions}, volume 382.
\newblock John Wiley \& Sons, 2007.

\bibitem[Baum(1972)]{baum1972equality}
L.~E. Baum.
\newblock An equality and associated maximization technique in statistical
  estimation for probabilistic functions of markov processes.
\newblock \emph{Inequalities}, 3:\penalty0 1--8, 1972.

\end{thebibliography}

\clearpage

\section*{Supplement: Experimental Details}

\setcounter{section}{1}

\subsection{Box2D Simulation: 2D Surface Pushing with Friction}

\vspace{0.25em}\noindent \textbf{Algorithmic Supervisor: } Our algorithmic supervisor runs an RRT Connect motion planner (from the Open Motion Planning Library \textsf{ompl}) at each time-step planning till the goal. The motion planning algorithm contains a single important hyper-parameter which is the maximum length of branches to add to the tree. We set this at 0.1 units (for a 20 x 20 unit space).

\begin{figure}[t]
    \centering
    \includegraphics[width=\textwidth]{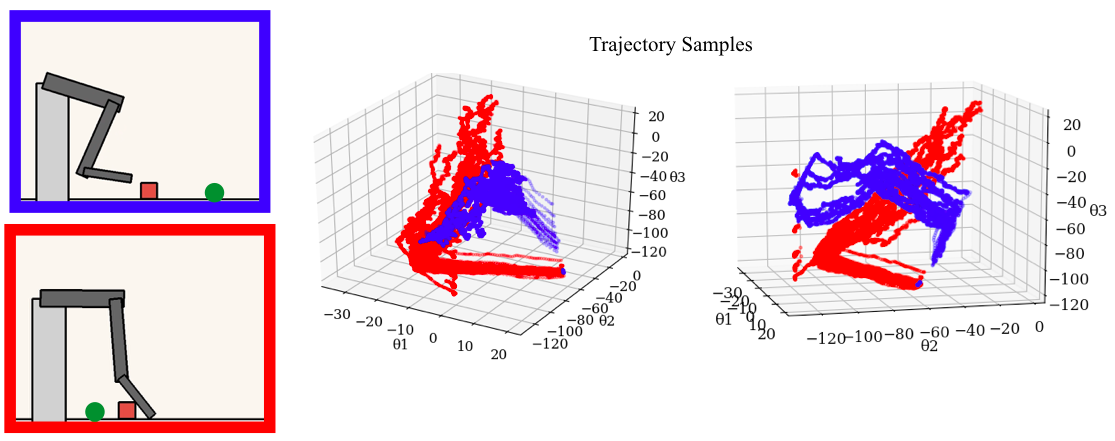}
    \caption{Due to the collision and contact constraints of the 2D Surface Pushing task, it leads to two geometrically distinct pushing skills, backhand and forehand, for goals on the left and on the right. Right: 50 sampled trajectories from a motion-planning supervisor. \label{fig:b2dcspace}}
\end{figure}

\vspace{0.25em}\noindent \textbf{Task Geometry: } We designed this experiment to illustrate how options can lead to more concise representations since they can specialize in different regions of the state-spaces.
Due to the geometry of the configuration space and the task, there are two classes of trajectories that are generated, when the goal is to the left or right of the arm.
The 50 sampled trajectories plotted in joint angle space in Figure \ref{fig:b2dcspace} are clearly separated into two distinct skills, backhand and forehand. 
Furthermore, these skills can be implemented by a locally affine policies in the joint angle space.

\vspace{0.25em}\noindent \textbf{Multi-Layer Perceptron Flat Policy: } One of the baseline policies is a multi-layer perceptron (MLP) policy which has a single ReLU hidden layer of 64 nodes.
This policy is implemented in Tensorflow and is trained with an ADAM optimizer with learning rate $1e-5$.

\vspace{0.25em}\noindent \textbf{\alg Policy 1: } In the first policy trained by \alg, we have a logistic regression meta-policy that selects from one of $k$ linear sub-policies. The linear sub-policies execute until a termination condition determined again by a logistic regression. This policy is implemented in Tensorflow and is trained with an ADAM optimizer with learning rate $1e-5$. For the linear hierarchy, we set \alg to discover 5 options, which is tuned using the cross-validation method described in the paper.

\vspace{0.25em}\noindent \textbf{\alg Policy 2: } In the second policy trained by \alg, we have a logistic regression meta-policy that selects from one of $k$ multi-layer perceptron sub-policies. 
As with the flat policy, it has a single ReLU hidden layer of 64 nodes.
The MLP sub-policies execute until a termination condition determined again by a logistic regression. This policy is implemented in Tensorflow and is trained with an ADAM optimizer with learning rate $1e-5$. For the MLP hierarchy, we set \alg to discover 2 options, which is tuned using the cross-validation method described in the paper.

\subsection{Image-Based Surface Pushing with Friction}
 Next, we consider the case where the sensor inputs are 640x480 images of the scene. 
 
 \vspace{0.25em}\noindent \textbf{\alg FF Policy: } First, we consider a neural network with a convolutional layer with 64 5x5 filters followed by a fully connected layer forming a feed-forward (FF) network. The high-level model only selects options and is parametrized by the same general architecture with an additional softmax layer after the fully connected layer. This means that the meta-control policy is a two-layer convolutional network whose output is a softmax categorical distribution over options. This policy is implemented in Tensorflow and is trained with an ADAM optimizer with learning rate $1e-5$.
 We used $k=2$ options for the FF policy.
 
 \vspace{0.25em}\noindent \textbf{\alg RNN Policy: } Next, we consider a neural network with a convolutional layer with 64 5x5 filters followed by an LSTM layer forming  a recurrent network (RNN). The high-level model only selects options and is parametrized by the same general architecture with an additional softmax layer after the LSTM. This policy is implemented in Tensorflow and is trained with a Momentum optimizer with learning rate $1e-4$ and momentum $4e-3$.
  We used $k=4$ options for the RNN policy.
  
  \begin{figure}[t]
    \centering
    \includegraphics[width=0.48\textwidth]{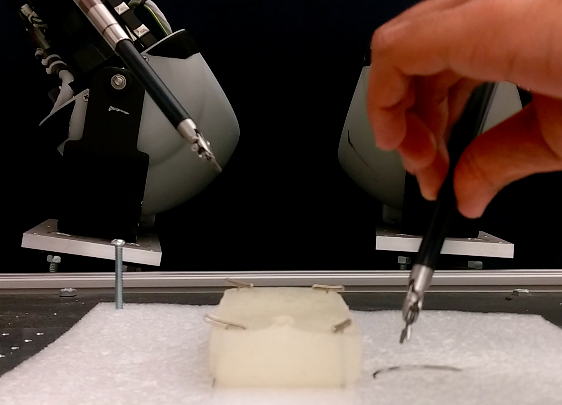}
    \includegraphics[width=0.455\textwidth]{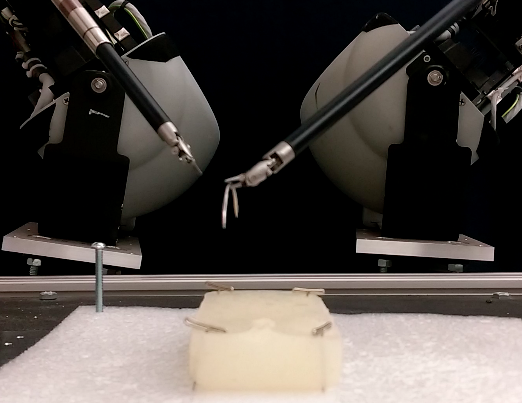}
    \caption{To collect demonstrations, we programmed an initial open-loop control policy. We observed the policy execute and interrupted the robot via keyboard input when adjustment was needed.  \label{fig:demo}}
\end{figure}
 
\subsection{Physical Experiment 1: Needle Insertion}

\vspace{0.5em} \noindent \textbf{Task Description: } The robot must grasp a 1mm diameter surgical needle, re-orient it parallel to a tissue phantom, and insert the needle into a tissue phantom.
The task is successful if the needle is inserted into a 1 cm diameter target region on the phantom.
In this task, the state-space is the current 6-DoF pose of the robot gripper and visual features that describe the estimated pose of the needle.
These features are derived from an image segmentation that masks the needle from the background and fits an ellipsoid to the resulting pixels. 
The principal axis of this 2D ellipsoid is a proxy for the pose of the needle.
The task runs for a fixed 15 time-steps and the policy must set the joint angles of the robot at each time-step.

\vspace{0.5em} \noindent \textbf{Robot Parameters: } The challenge is that the curved needle is sensitive to the way that it is grasped. Small changes in the needle's initial orientation can lead to large changes to the in-gripper pose of the needle due to deflection.
This deflection coupled with the inaccurate kinematics of the DVRK leads to very different trajectories to insert the needle properly.

The robotic setup includes a stereo endoscope camera located 650 mm above the 10cm x 10 cm workspace.
After registration, the dvrk has an RMSE kinematic error of 3.3 mm, and for reference, a gripper width of 1 cm.
In some regions of the state-space this error is even higher, with a 75\% percentile error of 4.7 mm.
The learning in this task couples a visual servoing policy to grasp the needle with the decision of which direction to orient the gripper after grasping.

\vspace{0.5em} \noindent \textbf{Demonstration Protocol: }
To collect demonstrations, we programmed an initial open-loop control policy. This policy traced out the basic desired robot motion avoiding collisions and respecting joint limits, and grasping at where it believed the needle was and an open-loop strategy to pin the needle in the phantom.
This was implemented by 15 joint angle way points which were interpolated by a motion planner.
We observed the policy execute and interrupted the robot via keyboard input when adjustment was needed.
This interruption triggered a clutching mechanism and we could 
kinesthetically adjusted the joints of the robot and pose of the gripper (but not the open-close state).
The adjustment was recorded as a delta in joint angle space which was propagated through the rest of the trajectory.
We collected 100 such demonstrations and images of these adjustements are visualized in image (Figure \ref{fig:demo}).

\begin{figure}
    \centering
    \includegraphics[width=\textwidth]{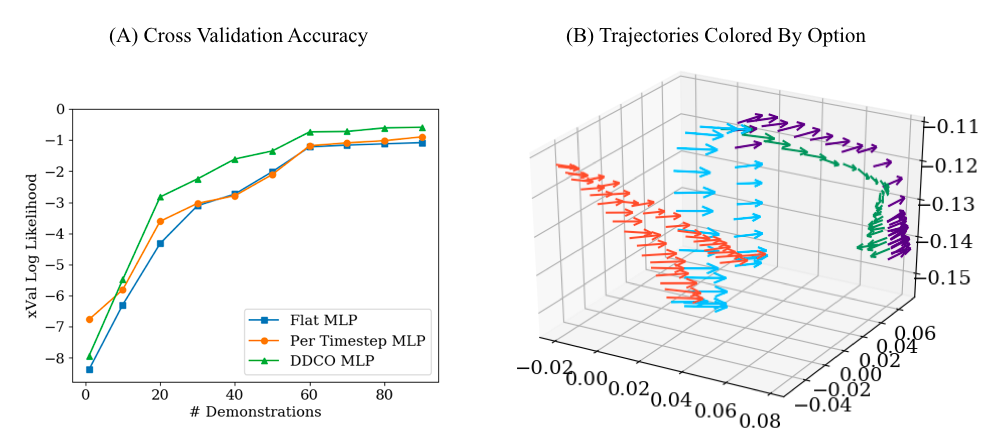}
    \caption{(A) We plot the cross validation likelihood of the different methods as a function of the number of demonstrations. (B) We visualize two representative trajectories (position and gripper orientation) color coded by the most likely option applied at that timestep. We find that the two trajectories have the same first two options but then differ in the final step due to the re-orientation of the gripper before insertion.  \label{fig:dvrkexp1}}
\end{figure}

\vspace{0.25em} \noindent \textbf{Learning the Parameters: } Figure \ref{fig:dvrkexp1}A plots the cross-validation log-likelihood as a function of the number of demonstrations.
We find that the hierarchical model has a higher likelihood than the alternatives---meaning that it more accurately explains the observed data and generalizes better to held out data.
At some points, the relative difference is over 30\%.
It, additionally, provides some interpretability to the learned policy.
Figure \ref{fig:dvrkexp1}B visualizes two representative trajectories.
We color code the trajectory based on the option active at each state (estimated by \alg).
The algorithm separates each trajectory into 3 segments: needle grasping, needle lifting, and needle orienting.
The two trajectories have the same first two options but differ in the orientation step.
One of the trajectories has to rotate in a different direction to orient the needle before insertion.

\vspace{0.25em} \noindent \textbf{Flat Policy: } One of the baseline policies is a multi-layer perceptron (MLP) policy which has a single ReLU hidden layer of 64 nodes.
This policy is implemented in Tensorflow and is trained with an ADAM optimizer with learning rate $1e-5$.

\vspace{0.25em} \noindent \textbf{Per-Timestep Policy: } Next, we consider a degenerate case of options where each policy executes for a single-timestep. We train 15 distinct multi-layer perceptron (MLP) policies each of which has a single ReLU hidden layer of 64 nodes.
Thes policies are implemented in Tensorflow and are trained with an ADAM optimizer with learning rate $1e-5$.

\vspace{0.25em} \noindent \textbf{\alg Policy: } \alg trains a hierarchical policy with 5 options. We considered a hierarchy where the meta policy is a multilayer perceptron with a softmax output that selects the appropriate option, and the options are parametrized by another multilayer perceptron with continuous outputs. Each of the MLP policies has a single ReLU hidden layer of 64 nodes.
Thes policies are implemented in Tensorflow and are trained with an ADAM optimizer with learning rate $1e-5$.

\vspace{0.25em} \noindent \textbf{Execution: } 
For each of the methods, we execute ten trials and report the success rate (successfully grasped and inserted the needle in the target region), and the accuracy.
The results are described in aggregate in the table below:

\begin{table}[ht!]\footnotesize
\centering
\label{my-label}
\begin{tabular}{l|l|l|l|l|}
             & Overall Success & Grasp Success & Insertion Success & Insertion Accuracy \\
             \hline
Open Loop    & 2/10   & 2/10  & 0/0 & $7 \pm 1$ mm                  \\
Behavioral Cloning & 3/10 & 6/10  & 3/6 &   $6 \pm 2$ mm     \\
Per Timestep & 6/10  & 7/10  & 6/7 &   $5 \pm 1$ mm \\ 
\alg & 8/10 & 10/10 & 8/10 &  $5 \pm 2$ mm  \\
\end{tabular}
\end{table}

\begin{figure}
    \centering
    \includegraphics[width=0.8\textwidth]{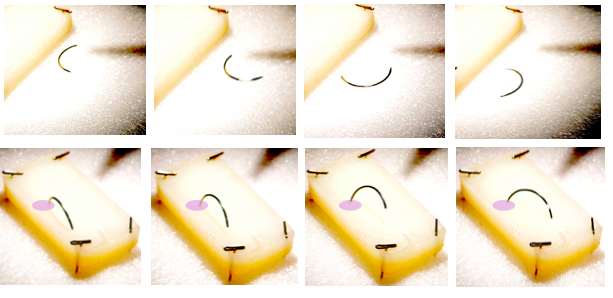}
    \caption{Illustration of the needle orientation and insertion task. Above are images illustrating the variance in the initial state, below are corresponding final states after executing \alg.  \label{fig:dvrkexp3-1}}
\end{figure}

\begin{figure}
    \centering
    \includegraphics[width=0.25\textwidth]{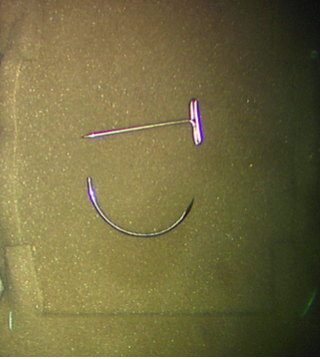}
    \includegraphics[width=0.4\textwidth]{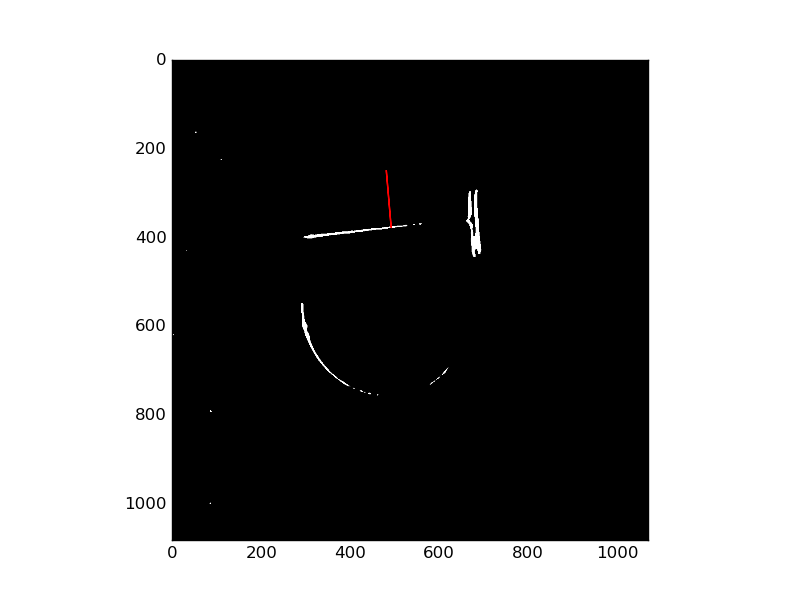}
    \caption{The endoscope image and a corresponding binary mask with a selected grasp. The arrow corresponds to the orientation of the gripper along the grasp axis.  \label{fig:dvrkim}}
\end{figure}

\subsection{Physical Experiment 2: Surgical Bin Picking}
\vspace{0.5em} \noindent \textbf{Task Description: } We consider a task with a more complicated high-level structure.
In this task, the robot is given a foam bin with 3 different types of needles (1mm-3mm in diameter) lying in a pile (5-8 needles in experiments).
The robot must extract a particular type of needle  and place it in the accept cup and place all others in a reject cup.
The task is successful if the entire foam bin is cleared.

\begin{figure}
    \centering
    \includegraphics[width=0.6\textwidth]{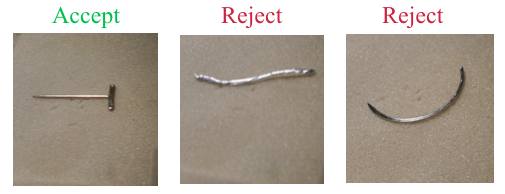}
    \caption{There are two bins, one accept and one reject bin. In the accept bin, we place dissection pins and place the suturing needles and the wires in the other. \label{fig:dvrkcats}}
\end{figure}

Figure \ref{fig:dvrkcats} shows representive objects for this task. We consider three types of ``needles'': dissection pins, suturing needles, and wires.
Dissection pins are placed in the accept cup and the other two are placed in the reject cup.

\vspace{0.5em} \noindent \textbf{Robot and State-Space Parameters: } As in the previous task, the task requires learning because the kinematics of the dvrk are such that the precision needed for grasping needles requires visual servoing.
However, even with visual servoing, failures are common due to the pile (grasps of 2, 3, 4 objects).
We would like to automatically learn recovery behaviors. 
In our robotic setup, there is an overhead endoscopic stereo camera, and it is located 650mm above the workspace.

To define the state-space for this task, we first generate binary images from the stereo images and apply a color-based segmentation to identify the needles (we call this feature \textsf{image}).
Then, we use a classifier derived from 40 hand-labeled images to identify possible grasp points to sample a candidate grasp ( left pixel value, right pixel value, and direction) (we call this feature \textsf{grasp}). 
These features are visualized in Figure \ref{fig:dvrkim}.
Additionally, there is the 6 DoF robot gripper pose and the open-close state of the gripper (we call this feature \textsf{kin}).
The state-space of the robot is (\textsf{kin}, \textsf{image}, \textsf{grasp}), and the action space for the robot is 6 joint angles and the gripper angle.
Each grasp, lift, move, and drop operation consists of 10 time steps of joint angle positions. The motion between the joint angles is performed using a SLURP-based motion planner.

\vspace{0.5em} \noindent \textbf{Demonstration Protocol: } As in the previous task, to collect demonstrations, we start with a hard-coded open-loop policy.
We roll this policy out and interrupt the policy when we anticipate a failure.
Then, we kinesthetically adjust the pose of the dvrk and it continues.
We collected 60 such demonstrations of fully clearing the bin filled with 3 to 8 needles each--corresponding to 450 individual grasps.
We also introduced a key that allows the robot to stop in place and drop it current grasped needle.
Recovery behaviors were triggered when the robot grasps no objects or more than one object.
Due to the kinesthetic corrections, a very high percentage of the attempted grasps (94\%) grasped at least one object. 
Of the successful grasps, when 5 objects are in the pile 32\% grasps picked up 2 objects, 14\% picked up 3 objects, and 0.5\% picked up 4.
In recovery, the gripper is opened and the epsiode ends leaving the arm in place.
The next grasping trial starts from this point.

\begin{figure} [t]
    \includegraphics[width=0.48\textwidth]{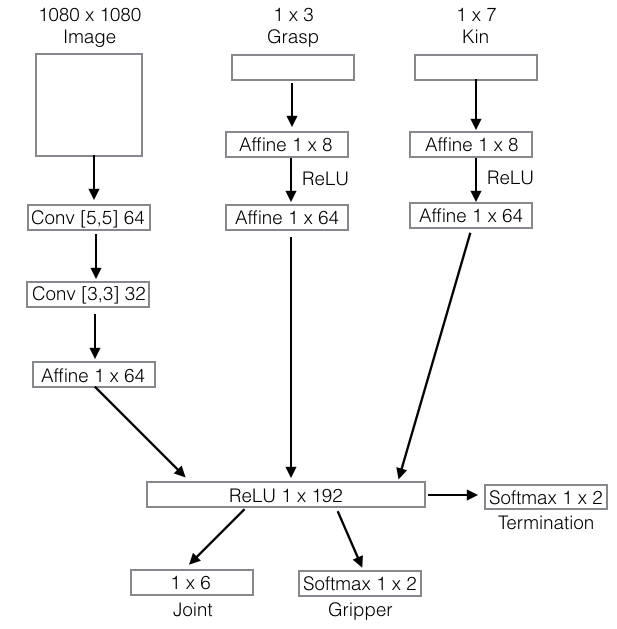}
    \includegraphics[width=0.48\textwidth]{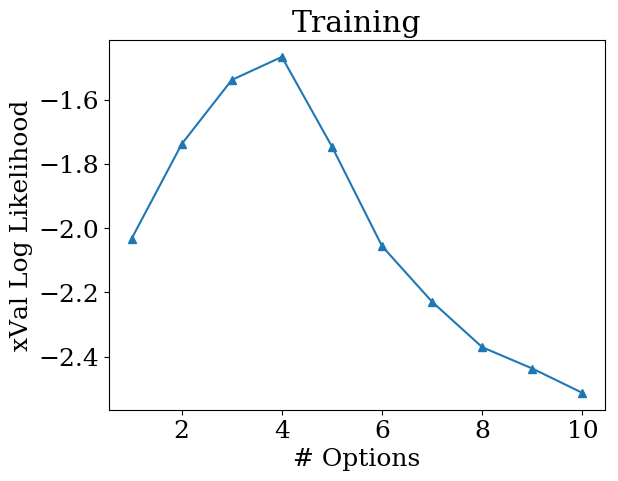}
    \caption{We use the above neural network to parametrize each of the four options learned by \alg. In this network, there are three inputs the a binary image, a candidate grasp, and kinematics. These inputs are processed through individual neural net branches and aggregated into a single output layer. This output layer sets the joint angles of the robot and the gripper state (open or closed). Option termination is also determined from this output. \label{fig:dvrknet}}
\end{figure}

\vspace{0.25em} \noindent \textbf{Policy Parametrization: } We apply \alg to the collected demonstrations with the policy parametrization described in Figure \ref{fig:dvrknet}. In this network, there are three inputs the a binary image, a candidate grasp, and kinematics. These inputs are processed through individual neural net branches and aggregated into a single output layer. This output layer sets the joint angles of the robot and the gripper state (open or closed). Option termination is also determined from this output. Using the cross-validation technique described in the paper, we identify 4 options (Figure \ref{fig:dvrknet}). The high-level policy has the same parametrization but after the output layer there is a softmax operation that selects a lower-level option.  

\begin{figure} [t]
    \includegraphics[width=\textwidth]{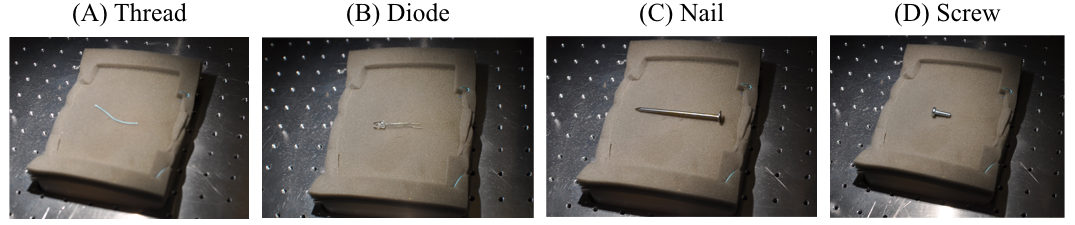}
    \caption{We evaluated the generalization of the learned policy on a small set of unseen objects. This was to understand what features of the object binary mask is used to determine behaviors.\label{fig:dvrkgen}}
\end{figure}

\vspace{0.25em} \noindent \textbf{Coordinate Systems: } We also experimented with different action space representations to see if that had an effect on single object grasps and categorizations. We trained alternative policies with the collected dataset where instead of predicting joint angles, we alternatively predicted 6 DoF end-effector poses with a binary gripper open-close state , and end-effector poses represented a rotation/translation matrix with a binary gripper open-close state. 
We found that the joint angle representation was the most effective. 
In particular, we found that for the grasping part of the task, a policy that controlled the robot in terms of tooltip poses was unreliable.

\begin{table}[ht!]\footnotesize
\centering
\label{my-label}
\begin{tabular}{l|l|l|l|l|}
       & Items & Successful Grasp & Successful Recovery & Successful Categorizations \\
       \hline
Joint Angle & 8     & 7/8         & 2/2    & 7/7                         \\
Tooltip Pose  & 8     & 3/8         & 5/5    & 3/3                             \\
Rotation   & 8     & 2/8          & 0/8    & 0                     
\end{tabular}
\end{table}
\begin{figure} [t]
\centering
    \includegraphics[width=0.8\textwidth]{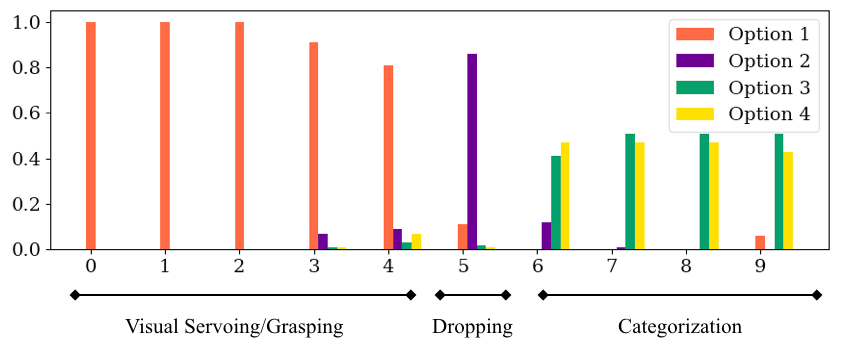}
    \caption{We plot the time distribution of the options selected by the high-level policy. The x-axis represents time, and the bars represent the probability mass assigned to each option. We find that the structure of the option aligns with key phases in the task such as servoing to the needle, grasping it, and categorizing. \label{fig:dvrkhl}}
\end{figure}

\vspace{0.25em} \noindent \textbf{Interpreting Learned Options: }
We additionally analyzed the learned options to see if there was an interpretable structure.
We examined the collected demonstrations and looked at the segmentation structure.
We average over the 60 trials the probability for the high-level policy to choose each option in each of first 10 time-steps during training (Figure \ref{fig:dvrkhl}). We find that the options indeed cluster visited states and segment them in alignment with key phases in the task, such as servoing to the needle, grasping it, dropping it if necessary, and categorizing it into the accept cup. 

\vspace{0.25em} \noindent \textbf{Full Break Down of Experimental Results: }
In 10 trials, 7 out of 10 were successful (Table \ref{big-experiment}). The main failure mode was unsuccessful grasping (defined as either no needles or multiple needles). As the piles were cleared and became sparser, the robot's grasping policy became somewhat brittle. The grasp success rate was 66\% on 99 attempted grasps. In contrast, we rarely observed failures at the other aspects of the task. Of the grasps that failed, nearly 34\% were due to grasping multiple objects.

\begin{table*}[ht!]\footnotesize
\centering
\label{big-experiment}
\begin{tabular}{llllll}
\rowcolor[HTML]{000000} 
{\color[HTML]{FFFFFF} Trial \#} & {\color[HTML]{FFFFFF} \# of Needles} & {\color[HTML]{FFFFFF} \# Needles Cleared} & {\color[HTML]{FFFFFF} Grasping} & {\color[HTML]{FFFFFF} Recovery} & {\color[HTML]{FFFFFF} Categorization} \\
1                               & 6                                    & {\color[HTML]{32CB00} \textbf{6}}         & 6/9                             & 3/3                             & 6/6                                   \\
2                               & 8                                    & 6                                         & 7/12                            & 5/6                             & 6/7                                   \\
3                               & 7                                    & {\color[HTML]{32CB00} \textbf{7}}         & 7/8                             & 1/1                             & 7/7                                   \\
4                               & 6                                    & {\color[HTML]{32CB00} \textbf{6}}         & 6/10                            & 4/4                             & 6/6                                   \\
5                               & 7                                    & 6                                         & 6/11                            & 5/5                             & 6/6                                   \\
6                               & 8                                    & {\color[HTML]{32CB00} \textbf{8}}         & 8/13                            & 5/5                             & 8/8                                   \\
7                               & 6                                    & 5                                         & 5/9                             & 4/4                             & 5/5                                   \\
8                               & 7                                    & {\color[HTML]{32CB00} \textbf{7}}         & 7/10                            & 3/3                             & 7/7                                   \\
9                               & 8                                    & {\color[HTML]{32CB00} \textbf{8}}         & 8/10                            & 2/2                             & 8/8                                   \\
10                              & 6                                    & {\color[HTML]{32CB00} \textbf{6}}         & 6/7                             & 1/1                             & 6/6                                   \\
Totals                          & 69                                   & Success: 70\%                             & Success: 66\%                   & Success: 97\%                   & Success: 98\%                        
\end{tabular}
\end{table*}

\vspace{0.25em} \noindent \textbf{Generalization to unseen objects: } We also evaluated the learned policy on a few unseen objects (but similarly sized and colored) to show that there is some level of generalization in the learning. We tried out four novel objects and evaluated what the learned policy did for each (Figure \ref{fig:dvrkgen}). For each of the novel objects we tried out 10 grasps in random locations and orientations in the bin (without any others). We evaluated the grasp success and whether it was categorized consistently (i.e., does the learned policy consistently think it is a pin or a needle).

We found that the diode was consistently grasped and categorized as a dissection pin. We conjecture this is because of its head and thin metallic wires. On the other hand, the screw and the thread were categorized in the reject cup. For 8/10 of the successful grasps, the nail was categorized as a failure mode. We conjucuture that since it is the large object it looks similar to the two object grasps seen in the demonstrations.

\begin{table}[ht!]\footnotesize
\centering
\label{my-label}
\begin{tabular}{l|l|l|l|l|l}
       & Grasps & Successful & Drop & Categorize Accept & Categorize Reject \\
       \hline
Thread & 10     & 10         & 1    & 0              & 9                 \\
Diode  & 10     & 10         & 0    & 10             & 0                 \\
Nail   & 10     & 8          & 8    & 0              & 0                 \\
Screw  & 10     & 4          & 0    & 0              & 4                
\end{tabular}
\end{table}

\subsection{Action Augmentation: Image-based Pushing}
We run the same experiment as described in the paper but on the image state-space instead of the low dimensional state.
The sensor inputs are 640x480 images of the scene and the task is the Box2D pushing task. 
 
 \vspace{0.25em}\noindent \textbf{\alg FF Policy: } First, we consider a neural network with a convolutional layer with 64 5x5 filters followed by a fully connected layer forming a feed-forward (FF) network. The high-level model only selects options and is parametrized by the same general architecture with an additional softmax layer after the fully connected layer. This means that the meta-control policy is a two-layer convolutional network whose output is a softmax categorical distribution over options. This policy is implemented in Tensorflow and is trained with an ADAM optimizer with learning rate $1e-5$.
 We used $k=2$ options for the FF policy.
 
 \vspace{0.25em}\noindent \textbf{\alg RNN Policy: } Next, we consider a neural network with a convolutional layer with 64 5x5 filters followed by an LSTM layer forming  a recurrent network (RNN). The high-level model only selects options and is parametrized by the same general architecture with an additional softmax layer after the LSTM. This policy is implemented in Tensorflow and is trained with a Momentum optimizer with learning rate $1e-4$ and momentum $4e-3$.
  We used $k=4$ options for the RNN policy.

\begin{figure} [t]
\centering
    \includegraphics[width=0.4\textwidth]{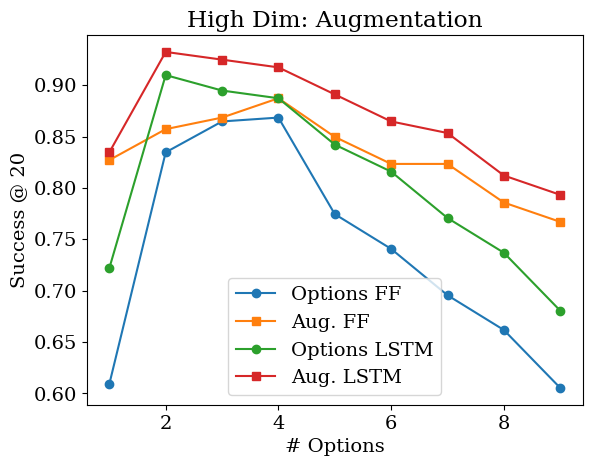}
    \caption{Augmenting the action space instead of selecting only options with the meta-policy leads attenuates the losses when selecting too few or too many options. \label{fig:augimages}}
\end{figure}

We test two different architectures for the output layer of this high-level policy: either a softmax categorical distribution selecting an option, or the hybrid softmax/continuous distribution output described in~\sref{hybrid}.
Figure \ref{fig:augimages} describes the empirical estimation, through policy rollouts, of the success rate as a function of the number of options, and of the fraction of time that the high-level policy applies physical control.
When the options are too few to provide skills that are useful throughout the state space, physical controls can be selected instead to compensate in states where no option would perform well. 

\section*{Supplement: Additional Experiments on Algorithm Stability}
\setcounter{section}{2}

We also ran a number of experiments to evaluate the stability of the solutions found with \alg. In other words, given different initializations, how consistent are the learned options in terms of quality and structure.

\subsection{Instability Modes}
When real perceptual data is used, if all of the low-level policies initialize randomly the forward-backward estimates needed for the Expectation-Gradient will be very poorly conditioned where there is an extremely low likelihood assigned to any particular observation.
We encountered the following degenerate cases in \alg:

\vspace{0.25em} \textbf{High-Fitting: } When the high-level policies can invoke both actions and options, one degenerate solution is ``high fitting''--where the meta-policy disregards all of the options. 

\vspace{0.25em} \textbf{Imbalance: } If the options are highly expressive, another degenerate case is when the options are imbalanced, i.e., one option dominates over a large portion of the state-space.

\vspace{0.25em} \textbf{Redundancy: } Options that essentially perform the same sub-task might be learned.

We find that these concerns can be greatly mitigated with more intelligent initialization and layer-wise training.   

\subsection{Vector Quantization For Initialization}
One challenge with \alg is initialization.
When real perceptual data is used, if all of the low-level policies initialize randomly the forward-backward estimates needed for the Expectation-Gradient will be poorly conditioned where there is an extremely low likelihood assigned to any particular observation.
The EG algorithm relies on a segment-cluster-imitate loop, where initial policy guesses are used to segment the data based on which policy best explains the given time-step, then the segments are clustered, and the policies are updated.
In a continuous control space, a randomly initialized policy may not explain any of the observed data well.
This means the small differences in initialization can lead to large changes in the learned hierarchy.

We found that a necessary pre-processing step was a variant of vector quantization, originally proposed for problems in speech recognition. 
We first cluster the state observations using a \textsf{k-means} clustering and train $k$ behavioral cloning policies for each of the clusters.
We use these $k$ policies as the initialization for the EG iterations.
Unlike the random initialization, this means that the initial low level policies will demonstrate some preference for actions in different parts of the state-space.
We set $k$ to be the same as the $k$ set for the number of options, and use the same optimization parameters.

\subsection{Layer-wise Hierachy Training}
We also found that layer-wise training of the hierarchy greatly reduced the likelihood of a degenerate solution.
While, at least in principle, one can train 
When the meta-policy is very expressive and the options are initialized poorly, sometimes the learned solution can degenerate to excessively using the meta-policy (high-fitting). 
We can avoid this problem by using a simplified parametrization for the meta-control policy $\eta_d$ used when discovering the low-level options. For example, we can fix a uniform meta-control policy that chooses each option with probability $\nicefrac 1 {k}$. Now, once these low-level options are discovered, then we can augment the action space with the options and train the meta-policy.

This same algorithm can recursively proceed to deep hierarchies from the lowest level upward: level-$d$ options can invoke already-discovered lower-level options; and are discovered in the context of a simplified level-$d$ meta-control policy, decoupled from higher-level complexity.
Perhaps counter-intuitively, this layer-wise training does not sacrifice too much during option discovery, and in fact, initial results seem to indicate that it improves the stability of the algorithm.
An informative meta-control policy would serve as a prior on the assignment of demonstration segments to the options that generated them, but with sufficient data this assignment can also be inferred from the low-level model, purely based on the likelihood of each segment to be generated by each option.

\begin{figure} [t]
\centering
    \includegraphics[width=0.3\textwidth]{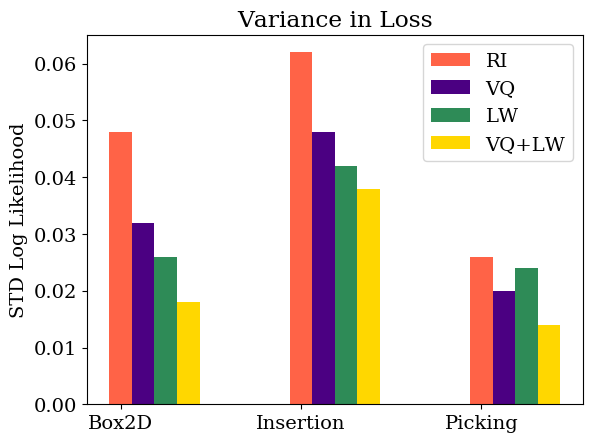}
    \caption{This experiment measures the variance in log likelihood over 10 different runs of \alg with different training and initialization strategies. Vector Quantization (VQ) and Layer-Wise training (LW) help stabilize the solution. \label{fig:exp81}}
\end{figure}

\begin{figure} [t]
\centering
    \includegraphics[width=0.3\textwidth]{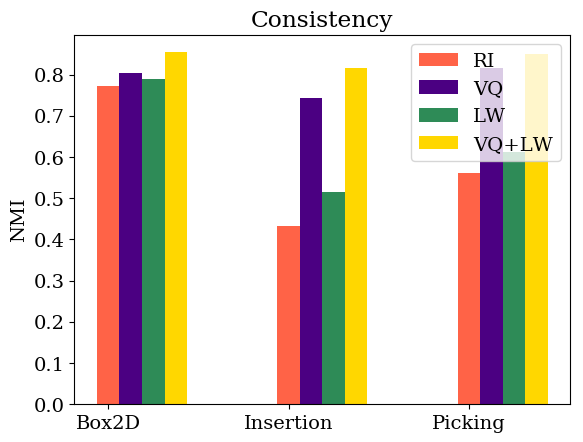}
    \caption{We measure the average normalized mutual information between multiple runs of \alg. This measures the consistency of the solutions across intializations. \label{fig:exp82}}
\end{figure}

\begin{figure} [t]
\centering
    \includegraphics[width=0.3\textwidth]{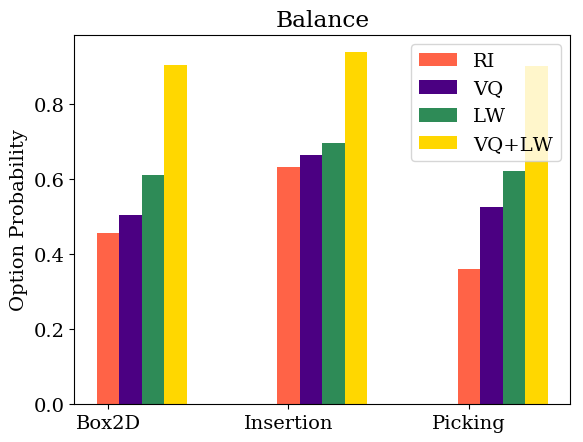}
    \caption{We measure probability that an option is selected by the high-level policy. \label{fig:exp83}}
\end{figure}

\subsection{Results: Stability of \alg}
In the first experiment, we measure the variance in log likelihood over 10 different runs of \alg with different training and initialization strategies (Figure \ref{fig:exp81}). 
We use the entire datasets presented in the paper and use the FF architecture for the Box2D experiments.
Vector Quantization (VQ) and Layer-Wise training (LW) help stabilize the solution and greatly reduce the variance of the algorithm.
This reduction in variance is over 50\% in the Box2D experiments, and a little more modest for the real data.

In the next experiment, we measure the consistency of the solutions found by \alg (Figure \ref{fig:exp82}).
For each of the demonstration trajectories, we annotate the time-step by the most likely option.
One can view this annotation as a hard clustering of the state-action tuples.
Then, we measure the average normalized mutual information (NMI) between all pairs of the 10 different runs of \alg.
NMI is a measure of how aligned two clusterings are between 0 and 1, where 1 indicates perfect alignment.
As with the likelihood, Vector Quantization (VQ) and Layer-Wise training (LW) significantly improve the consistency of the algorithm.

In the last experiment, we measure the symptoms of the ``high-fitting'' problem (Figure \ref{fig:exp83}).
We plot the probability that the high-level policy selects an option.
In some sense, this measures how much control the high-level policy delegates to the options.
Surprisingly, VQ and LW have an impact on this.
Hierarchies trained with VQ and LW have a greater reliance on the options.

\begin{figure} [t]
\centering
    \includegraphics[width=0.4\textwidth]{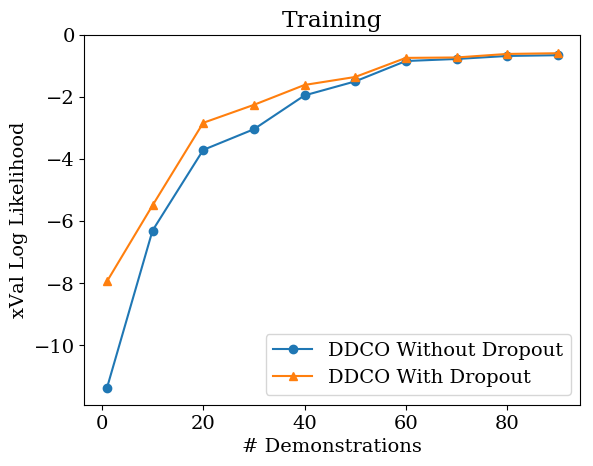}
    \caption{Dropout seems to have a substantial effect on improving cross-validation accuracy on a hold out set. \label{fig:exp9}}
\end{figure}

\subsection{Results: Dropout}
For the real datasets, we leverage a technique called dropout, which has been widely used in neural network training to prevent overfitting.
Dropout is a technique that randomly removes a unit from
the network along with all its connections.
We found that this technique improved performance when the datasets were small. We set the dropout parameter to 0.5 and measured the performance on a hold out set.
We ran an experiment on the needle insertion task dataset, where we measured the cross-validation accuracy on held out data with and without dropout (Figure \ref{fig:exp9}).
Dropout seems to have a substantial effect on improving cross-validation accuracy on a hold out set.

\end{document}